\documentclass[11pt]{article}
\usepackage{coling2020}
\usepackage{times}
\usepackage{url}
\usepackage{latexsym}
\usepackage{centernot}

\usepackage{booktabs} 
\usepackage{graphicx}
\usepackage{url}
\usepackage{mathtools}
\usepackage{xcolor}
\usepackage{comment}
\usepackage{makecell}
\usepackage[labelfont=bf]{caption}
\usepackage{subcaption}

\usepackage[ruled,vlined,linesnumbered]{algorithm2e}
\makeatletter
\def\BState{\State\hskip-\ALG@thistlm}
\makeatother
\usepackage{multirow}
\usepackage{setspace}
\usepackage[bottom]{footmisc}
\usepackage{graphicx}
\usepackage{booktabs}
\usepackage{makecell}
\usepackage{array}
\usepackage{bm}
\usepackage{balance}
\usepackage{amsmath}
\usepackage{colortbl}
\usepackage{amssymb}
\usepackage{tcolorbox}
\usepackage{setspace}
\usepackage{soul}
\usepackage{wrapfig}
\usepackage[export]{adjustbox}

\definecolor{Gray}{gray}{0.8}
\definecolor{Green}{RGB}{0, 100, 0}
\definecolor{LightBlue}{RGB}{180, 196, 222}
\definecolor{lilac}{RGB}{221, 200, 238}
\definecolor{paleyellow}{RGB}{255, 255, 161}
\newcommand*\samethanks[1][\value{footnote}]{\footnotemark[#1]} 

\DeclareRobustCommand{\hlpink}[1]{{\sethlcolor{paleyellow}\hl{#1}}}

\newcolumntype{a}{>{\columncolor{Gray}}c}
\colingfinalcopy 

\title{Interpretable Multi-Headed Attention for\\ Abstractive Summarization at Controllable Lengths}

\author{Ritesh Sarkhel\thanks{~~~The first two authors contributed equally to this work} \\
   \\
   \\
   \\
   \\\And
  
  Moniba Keymanesh\samethanks  \\
  \hspace{4cm}Department of Computer Science and Engineering \\
  \hspace{4.2cm}The Ohio State University \\
  \hspace{4.8cm}{\tt sarkhel.5,keymanesh.1,nandi.9,parthasarathy.2@osu.edu} \\\And
  
  Arnab Nandi \\
   \\
   \\
   \\
   \\\And
   
  Srinivasan Parthasarathy \\
   \\
   \\
   \\
   \\  }

\date{}

\begin{document}
\maketitle
\pagenumbering{gobble}
\vspace{-0.4cm}
\begin{abstract}

Abstractive summarization at controllable lengths is a challenging task in natural language processing. It is even more challenging for domains where limited training data is available or scenarios in which the length of the summary is not known beforehand. At the same time, when it comes to trusting machine-generated summaries, explaining how a summary was constructed in human-understandable terms may be critical. We propose {Multi-level Summarizer}~(\texttt{MLS}), a supervised method to construct abstractive summaries of a text document at controllable lengths. The key enabler of our method is an interpretable multi-headed attention mechanism that computes attention distribution over an input document using an array of timestep independent semantic kernels. Each kernel optimizes a human-interpretable syntactic or semantic property. Exhaustive experiments on two low-resource datasets in English language show that \texttt{MLS} outperforms strong baselines by up to 14.70\% in the METEOR score. Human evaluation of the summaries also suggests that they capture the key concepts of the document at various length-budgets.\looseness=-1\blfootnote{~This work is licensed under a Creative Commons Attribution 4.0 International License. License details: http://creativecommons.org/licenses/by/4.0/}

\end{abstract}

\section{Introduction}
Great progress has been made in recent years on abstractive summarization of text documents. Among existing works, sequence-to-sequence networks with attention~\cite{gehring2017convolutional,liu2018toward} have been one of the clear front-runners. Being able to constrain the length of a summary while preserving its desirable properties has many real-world applications. One such application is content optimization for variable screen-sizes. Online content creators such as news portals, blogs, and advertisement agencies with audiences on multiple platforms customize their content based on display-area for best experience. However, there has not been much work on summarization at controllable lengths until recently. High variance in screen-sizes often require extensive human supervision to perform these modifications. As most sequence-to-sequence networks~\cite{rush2015neural,nallapati2016abstractive} do not enforce the length of a summary, for scenarios as mentioned above, one may need to employ an ensemble of networks to cover all possible lengths. There are two major challenges in following this approach for real-world applications. First, training sequence-to-sequence networks is a resource-intensive task~\cite{strubell2019energy}. To train a network for generating summaries budgeted at length~$b$, we need a parallel corpus of text documents and their gold-standard summaries at length~$b$. Constructing a large enough corpus with summaries budgeted at~$b, \forall \hspace{0.02cm}b\in (0,1)$ may not be possible and/or cost-efficient for a number of domains. This is one of the main reasons why most existing works on abstractive summarization evaluate their model on large-scale news corpus datasets~\cite{nallapati2016abstractive,hermann2015teaching}, leaving out a number of important but low-resource domains~\cite{magooda2020abstractive,parida-motlicek-2019-abstract} where the number of available training documents is limited. Second, the range of possible length-budgets~$\mathcal{R}(b)$ may not always be known beforehand. In many scenarios, it can be known as late as during run-time. Therefore, we formalize the summarization task addressed in this paper as follows.\looseness=-1\vspace{0.05cm} 

\textbf{Problem Definition: }Given a document~$S$ of length~$N$ (tokens) and a maximum token budget of~$b$, we aim to construct an abstractive summary~$s_b$ that satisfies the following conditions, \textbf{\texttt{C1:}} information redundancy is minimized in~$s_b$; \textbf{\texttt{C2:}} coverage of the major topics of~$S$ is maximized in~$s_b$; \textbf{\texttt{C3:}} length of $s_b$ is maximal within the specified budget~$b$ without adversely affecting the conditions~\texttt{C1} and~\texttt{C2} i.e., ${|s_b|}\leq b$ \& $\nexists\hspace{0.05cm}s_{c}$ such that ${|s_{b}|}<{|s_{c}|}\leq b$. \texttt{C1} and \texttt{C2} ensure that the properties of a high-quality summary is preserved in~$s_b$, whereas \texttt{C3} ensures that~$s_b$ is the largest possible summary that can be constructed within budget~$b$ without compromising its quality. Note that \texttt{C1} and \texttt{C2} are seemingly contradictory to each other as the length of the summary increases. Our goal is to find the optimal tradeoff.\looseness=-1

\begin{figure}[t]
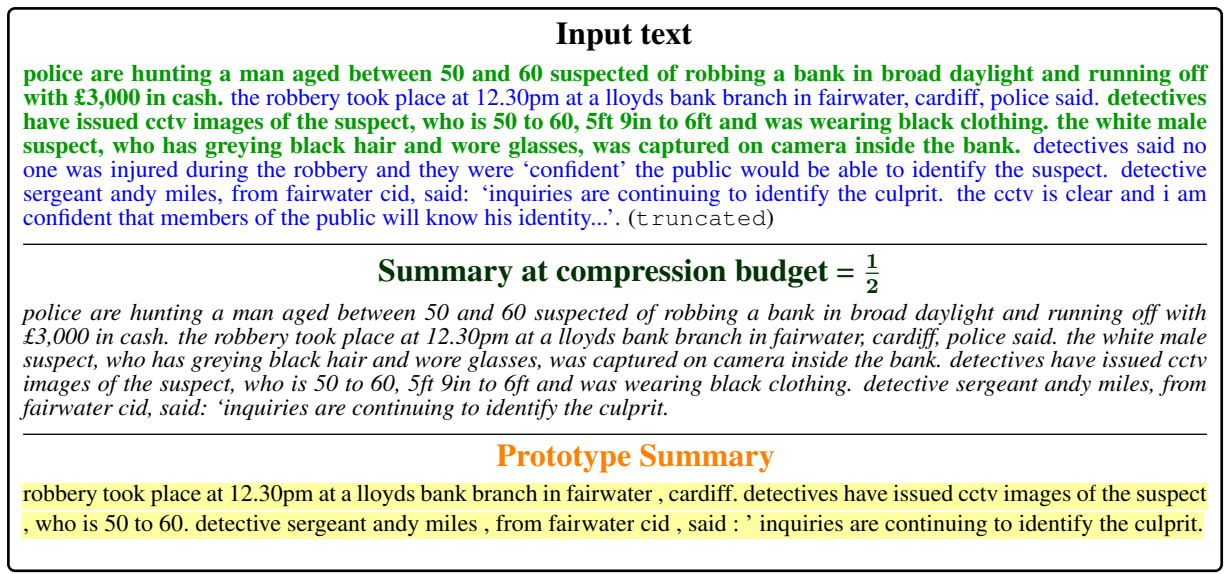

\begin{tcolorbox}[colback=white,colframe=black,boxrule=1pt,arc=2pt,boxsep=0pt,left=5pt,right=5pt,top=5pt,bottom=2pt]

\begin{spacing}{0.7}
\begin{small}
\hspace{0.45\linewidth}{\large{\textcolor{black}{{\textbf{Input text}}}}}\vspace{0.15cm}


\textbf{\textcolor{green!60!black}{police are hunting a man aged between 50 and 60 suspected of robbing a bank in broad daylight and running off with £3,000 in cash.}} \textcolor{blue}{the robbery took place at 12.30pm at a lloyds bank branch in fairwater, cardiff, police said.} \textbf{\textcolor{green!60!black}{{detectives have issued cctv images of the suspect, who is 50 to 60}, 5ft 9in to 6ft and was wearing black clothing. the white male suspect, who has greying black hair and wore glasses, was captured on camera inside the bank.}} \textcolor{blue}{detectives said no one was injured during the robbery and they were `confident' the public would be able to identify the suspect. {detective sergeant andy miles, from fairwater cid, said: `inquiries are continuing to identify the culprit.} the cctv is clear and i am confident that members of the public will know his identity...'.} (\texttt{truncated})\looseness=-1\vspace{0.05cm}

\noindent\rule[0.25ex]{\linewidth}{0.5pt}\vspace{0.05cm}

\hspace{0.3\textwidth}{\large{\textcolor{green!20!black}{\textbf{Summary at compression budget} = $\bm{\frac{1}{2}}$}}}\vspace{0.1cm}


\textit{police are hunting a man aged between 50 and 60 suspected of robbing a bank in broad daylight and running off with £3,000 in cash. the robbery took place at 12.30pm at a lloyds bank branch in fairwater, cardiff, police said. the white male suspect, who has greying black hair and wore glasses, was captured on camera inside the bank. detectives have issued cctv images of the suspect, who is 50 to 60, 5ft 9in to 6ft and was wearing black clothing. detective sergeant andy miles, from fairwater cid, said: `inquiries are continuing to identify the culprit.}\looseness=-1\vspace{0.05cm}

\noindent\rule[0.25ex]{\linewidth}{0.5pt}\vspace{0.05cm}

\hspace{0.4\textwidth}\textcolor{red!50!yellow}{\large{\textbf{Prototype Summary}}}\vspace{0.1cm}


\hlpink{robbery took place at 12.30pm at a lloyds bank branch in fairwater , cardiff. detectives have issued cctv images of the suspect , who is 50 to 60. detective sergeant andy miles , from fairwater cid , said : ' inquiries are continuing to identify the culprit.}\looseness=-1

\end{small}
\end{spacing}
\end{tcolorbox}\vspace{-0.2cm}
\begin{small}

\caption{\texttt{MLS} \textit{expands the highlighted sentences in the prototype summary to the boldfaced tokens in the input text to construct a summary budgeted at half-length of the input text \label{fig:mls_example}}\looseness=-1}
\end{small}
\vspace{-0.2cm}
\end{figure}

Early works on incremental summarization~\cite{buyukkokten2001seeing,yang2003fractal} leveraged structural tags supported by document markup languages to generate summaries at various lengths, thus imposing a serious constraint on the document formats~(e.g. XML, HTML) that come under the purview of such methods. Incremental sampling of sentences based on a salience score~\cite{otterbacher2006news,campana2009incremental} can partially solve this problem by constructing extractive summaries of the input document. We show in Section 3 that these sampling-based methods often fail to preserve the desirable properties of a high-quality summary. Among recent works, ~\cite{kikuchi2016controlling} were the first to propose a supervised method for controlling length during abstractive summarization. Their work was later extended by~\cite{fan2017controllable} who introduced the length of a summary as an input to the network. However, instead of exact input, they approximate the length to a predefined value-range, often failing to adhere to the allocated budget in a number of cases. \cite{liu2018controlling} address this issue by proposing a convolutional encoder-decoder network, introducing the desired summary length as an input to the initial state of the decoder. We compare and report its performance on two datasets in our experimental setup in Section 3.\looseness=-1 


Unfortunately, when it comes to {interpreting}\footnote{``the ability to explain or to present in understandable terms to a human'', \cite{doshi2017towards}} these models i.e. how the summaries came to be, the answer still remains illusive. Explaining how a machine-generated summary was constructed, has become a necessity under the newly introduced General Data Privacy Regulation Act~\cite{team2017eu}, especially for applications in enterprise~\cite{sarkhel2019visual,keymaneshtoward} and biomedical domain~\cite{moradi2018different,sarkhel2018nurses}. Some recent efforts have proposed using interpretable heatmaps~\cite{baan2019transformer} generated from the {attention distribution} over an input sequence for interpreting model behaviour. However, they are still quite limited~\cite{jain2019attention} in consistently explaining all aspects of a neural summarizer. This leaves a gap in the ongoing efforts~\cite{joint-parsing-summarization:2020,control-over-copying:2020} to generate abstractive summaries that are guided by human-interpretable semantic/syntactic qualities. Briefly, the main goal of attention mechanism in a encoder-decoder network is to assign a softmax score to every encoder hidden state (based on its relevance to the token being decoded) and amplify those that are assigned high scores through a weighted average. Source-target attention~\cite{nallapati2016abstractive} relies on another sequence for computing these scores, whereas self-attention~\cite{vaswani2017attention,paulus2018deep} operates over the elements in the current input sequence. A multi-headed attention mechanism allows a neural model to speed up training by enabling parallelization across timesteps. The number of operations in the computation of self-attention, however, scales quadratically with input length, making it a computationally expensive operation for long input sequences. Training such a network for a summarization task would require a large parallel corpus of input documents and their corresponding gold-standard summaries budgeted at~$b$. The role of some of the attention-heads during abstractive summarization is also not transparent~\cite{baan2019transformer}. To address these, we replace self-attention with a lightweight, interpretable alternative. Instead of projecting each input sequence multiple times\footnote{one time each to compute the query, key and value matrix~\cite{vaswani2017attention} from the input sequence} at every timestep, we encode an input sequence only once, using a timestep-independent kernel~($\Vec{Q}$) learned in an unsupervised or distantly supervised way from the input document. Each kernel has a human-interpretable syntactic/semantic role. Every attention-head in this multi-headed mechanism computes an attention distribution over the input sequence using a unique kernel~$\Vec{Q_i}$, recycling it at every timestep. 
Compared to self-attention, our proposed attention mechanism  scales linearly with the input sequence length and leverages significantly less number of trainable parameters. As we will show in Section 3, this allows us to train our network on limited training samples in low-resource datasets. 

\begin{wrapfigure}{l}{0.68\linewidth}
\vspace{-0.2cm}
\begin{center}
\includegraphics[width=\linewidth]{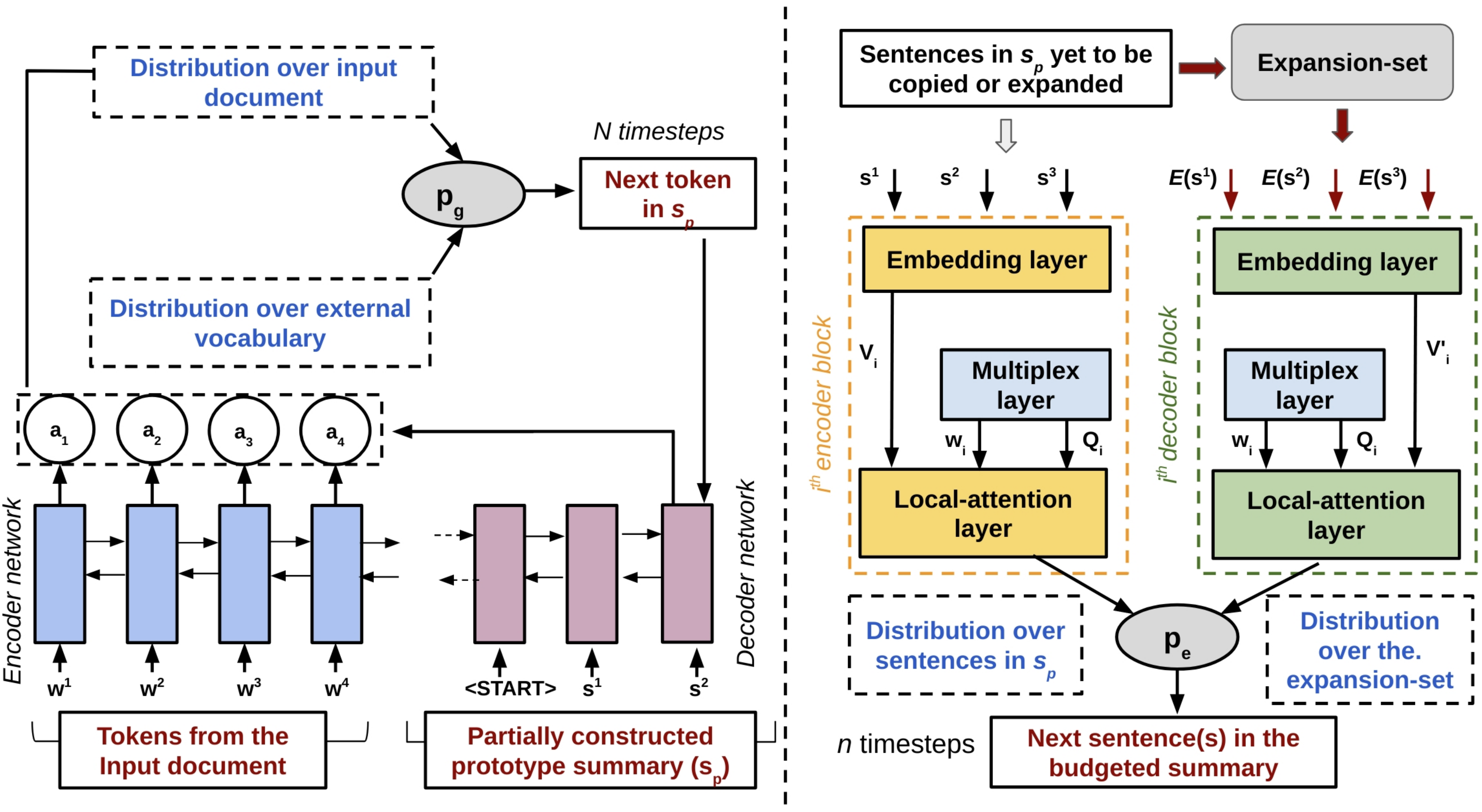}\label{fig:Picture2}
\end{center}
\begin{small}
\caption{\label{fig:Picture2}\textit{An overview of} \texttt{MLS} \textit{architecture. The PG-Network~(left) constructs a prototype summary~$s_p$ from the input document. The Pointer-Magnifier network~(right) constructs the length-constrained summary from~$s_p$ using interpretable sentence-level attention\looseness=-1}}\vspace{-0.2cm}
\end{small}
\end{wrapfigure}

We propose \texttt{MLS} -- a supervised method to generate abstractive summaries at arbitrary lengths in this paper. It computes a length-constrained summary~$s_b$ budgeted at length~$b$ by soft-switching between a copy and expand operation over a prototype summary~$s_p$ constructed from the document. The key enabler in this process is an {interpretable, multi-headed attention} mechanism. We develop a length-aware encoder-decoder network, called the \textit{Pointer-Magnifier} network that leverages this attention mechanism to construct summaries within a specified length. We train our network on limited training samples from two cross-domain datasets: the MSR-Narrative~\cite{ouyang2017crowd} and Thinking Machines dataset~\cite{thmch}. Exhaustive evaluation on a range of success metrics shows that \texttt{MLS} performs competitively or better against strong baseline methods. Subsequent human evaluation of summaries generated by \texttt{MLS} suggests that they accurately capture the main concepts of the input document. To summarize, some of the major contributions of this work are as follows:\looseness=-1\vspace{-0.1cm}


\begin{itemize}
    \item We propose \texttt{MLS}, a supervised approach to generate abstractive summaries of a text document at controllable lengths.
    
    \item We develop a length-aware encoder-decoder network that leverages an interpretable, multi-headed attention mechanism to construct length-constrained summaries. 
    
    \item Experimental results on two cross-domain datasets show that trained on limited training samples, \texttt{MLS} was able to generate summaries that are coherent and captured the key concepts of a document.\looseness=-1
\end{itemize}

\section{Proposed
Methodology}
{\texttt{MLS}} constructs a length-constrained summary of a document in two steps. First, it derives a prototype summary~$s_p$ from the document, covering its major concepts. Then, it expands or shortens it, depending on the length-budget to create the final summary. We employ a pair of encoder-decoder networks at both steps. For the first step, we extend the PG-network~\cite{see2017get}. We develop a length-aware encoder-decoder network for the second step. We describe both steps in greater detail in the following sections.\looseness=-1

\subsection{Generating the Prototype Summary}
We extend PG-Network by~\cite{see2017get} to construct the prototype summary~$s_p$ of a document. We tokenize the document and feed it to the encoder network sequentially. As the encoder hidden states are updated, the decoder network constructs the prototype summary one token at a time by soft-selecting between tokens in the input document and an external vocabulary. The decoding process is guided by an attention distribution\footnote{we closely followed the official implementation at: https://github.com/abisee/pointer-generator} computed over the input document and the external vocabulary. An overview of this network is shown in Fig~\ref{fig:Picture2}. We point the readers to the work by See et al. for more background on this network. An example prototype summary is shown in Figure~\ref{fig:mls_example}. Contrary to existing prototype-text guided summarization methods~\cite{liu2019automatic,saito2020length}, we do not specify the length of the prototype summary as an input of the network, rather infer it by outputting tokens until the EOS token is produced. We discuss the training and parameter settings of the network used in our experiments in Section 2.3. It is worth mentioning here that one of the main reasons to select the PG-Network as our architecture of choice for this step is due to its capability to construct a summary by looking up a learned language model. Other networks with similar capabilities can also be used, as this step has a transitive effect on the next phase of our approach.\looseness=-1

\subsection{Constructing the Length-Constrained Summary}
To construct a summary within length-budget~$b$, we develop the \textit{Pointer-Magnifier} network: a length-aware, interpretable, encoder-decoder network. An overview of the network is shown in~Fig.~\ref{fig:Picture2}. It consists of a multiplex layer, an encoder~(yellow rectangles) layer and a decoder~(green rectangles) layer. The encoder layer takes the prototype summary constructed in the previous step as input. The decoder layer outputs the final summary. We describe each layer in detail below.\looseness=-1\vspace{0.15cm}
 
\textbf{A.\hspace{0.2cm} The Multiplex Layer and Interpretable Kernels: }In an effort to build a transparent network, we embody three qualitative properties that are associated with a high-quality summary in our network. A high-quality summary, (1) maximizes the coverage of the \textit{major topics}~($\Phi_1$) and (2) \textit{keywords}~($\Phi_2$) appearing in the input document, while (3) minimizing the amount of \textit{redundant information}~($\Phi_3$). We encode each property using a semantic kernel~($\vec{Q_i}$), learned using an unsupervised or distantly supervised way from the input document itself. Every kernel plays a unique, human-interpretable syntactic/semantic role in constructing the final summary. One of the key components in this process is the multiplex layer~$\mathbb{M}$. Physically, it is a nested matrix of dimensions~$3 \times 3$ shared between the encoder and decoder layer. Each row in~$\mathbb{M}$ contains the following information: (a) a {distance-metric}~($dist_i$), (b) a scalar {value}~($w_i$), and (c) a semantic {kernel}~($\vec{Q_i}$), where $-1\leq w_i\leq 1, \forall i$ \& $\Sigma_i^3\hspace{0.05cm}w_i = 1$. During inference, each of these kernels measures the contribution of every sentence in the prototype summary towards optimizing one of the properties~$\Phi_i$, 1 $\leq i\leq$ 3, mentioned above. $w_i$ represents the relative weight assigned to the property~$\Phi_i$ in constructing the final summary. We compute the kernels as a preprocessing step.\looseness=-1\vspace{0.1cm} 

\textbf{Defining the Kernels: }To encode the property~$\phi_1$, we define~$\vec{Q_1}$ as a matrix of dimensions $3\times 300$, where each row of~$\vec{Q_1}$ represents one of the three most dominant topic vectors of the input document as a 300-dimensional vector. We use an unsupervised LDA-based model~\cite{blei2003latent} to derive these topic vectors. Symmetric KL-divergence is used as the distance metric~($dist_1$). Similarly, we encode the property~$\phi_2$ as a single dimensional vector~$\vec{Q_2}$ of length 50, where each vector component represents the relative frequency of one of the 50 most frequent keywords in the input document. We use RAKE~\cite{rose2010automatic}, a publicly available library to identify the keywords of a document. Symmetric KL-divergence is used as the distance metric~($dist_2$). Finally, we encode~$\phi_3$ as a matrix~$\vec{Q_3}$ of dimensions~$p\times 300$, where the~$i^{th}$ row of~$\vec{Q_3}$ represents an embedding of the~$i^{th}$ sentence in the input document. We compute the embedding vector of each sentence using a pretrained model~\cite{le2014distributed} on English Wikipedia corpus. Cosine similarity is used as the distance metric~($dist_3$). Our choice of unsupervised/distantly supervised kernels reflects our motivation~(see Section 1) to leverage a limited number of training samples from the experimental dataset to construct the final summary. We discuss the role played by each semantic kernel~($\vec{Q_i}$), distance metric~($dist_i$), and weight~($w_i$) in constructing the final summary from~$s_p$ in the following section.\looseness=-1\vspace{0.15cm} 

\begin{wrapfigure}{r}{0.65\linewidth}
  \begin{center}
  \vspace{-0.2cm}
    \includegraphics[width=\linewidth]{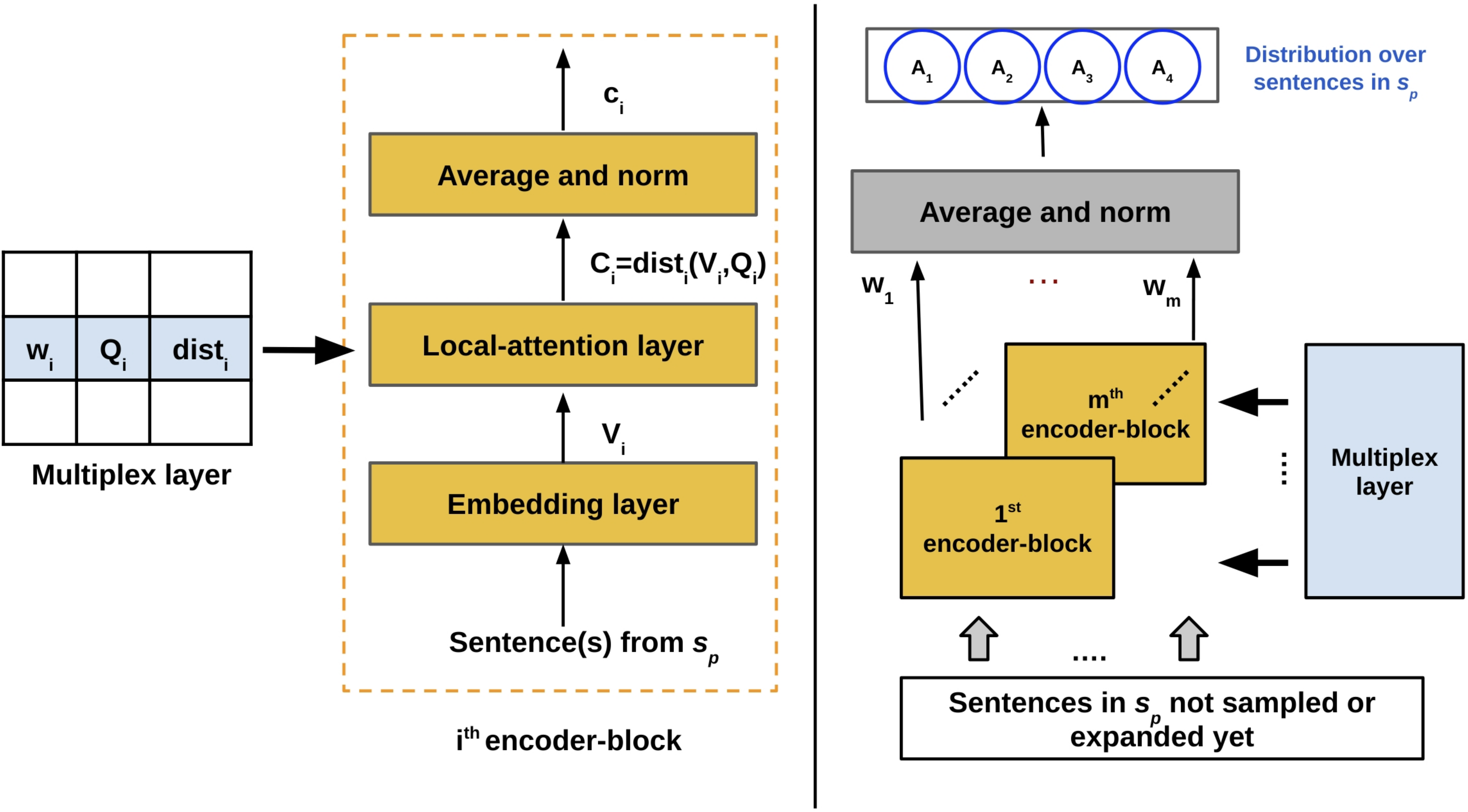}
  \end{center}\vspace{-0.25cm}
  \begin{small}
  \caption{\textit{The Encoder layer consists of 3 parallelly stacked encoder-blocks\looseness=-1}\label{fig:Picture3}}
  \end{small}
  \vspace{-0.3cm}
\end{wrapfigure}

\textbf{B.\hspace{0.2cm}The Encoder Layer: }The encoder layer consists of 3 parallelly stacked {encoder-blocks}. Each encoder-block~(see Fig.~\ref{fig:Picture3}) contains an \textit{embedding layer} and a \textit{local-attention layer}. At every timestep~$t$, a sentence from~$s_p$ is fed into the embedding layer of each of the three encoder-blocks. It computes a fixed-length embedding~($\vec{V_i}$) of the sentence and propagates it to the local-attention layer. Each encoder-block in our network is mapped to a unique triplet~($\vec{Q_i}$, $dist_i$, $w_i$) in the multiplex layer. To compute local-attention~($c_i$) attributed to a sentence in~$s_p$ by the~$i^{th}$ encoder-block, we embed it in the same semantic space as~$\vec{Q_i}$ and compute its distance from $\vec{Q_i}$ in that encoding space~(Eq.~1).\looseness=-1\vspace{-0.35cm}

\begin{gather}
\vec{C_{t,i}} = \frac{1}{r}\Sigma_{j=1}^r\hspace{0.05cm} dist_i(\vec{V_i},\hspace{0.05cm}\vec{Q_{i}}^{\,T}[j])\\
{c_i} = \frac{1}{n_i}\Sigma_{j=1}^{n_i} (\vec{C_{t,i}}[j])
\end{gather}

In Eq.~1, $\vec{Q_i}$ represents a kernel of dimensions~$r\times n_i$ and $\vec{V_i}$ represents an embedding vector of length~$n_i$. The embedding layer represents each sentence in~$s_p$ in the same encoding space as the kernel~$\vec{Q_i}$ associated with that block. We compute the local-attention~$c_i$ by taking a column-wise average of the distance-matrix~$\vec{C_{t,i}}$~(Eq.~2). The kernel~$\vec{Q_i}$ is reused for all the sentences fed to the $i^{th}$ encoder-block. The distribution $[c_1,c_2...]$ obtained this way is then normalized to derive the local-attention distribution~$\vec{C_i}$ over~$s_p$. The final attention distribution~($\vec{A}$) over~$s_p$ at timestep~$t$ is computed by normalizing the weighted average~(Eq.~3) of local-attention distributions computed by each attention-head.\looseness=-1\vspace{-0.5cm}

\begin{gather}
\vec{A^*} = \texttt{norm}(\frac{1}{m}\Sigma_i^m\hspace{0.05cm} (\vec{C_i}\cdot w_i))
\end{gather}\vspace{-0.4cm}

It is worth noting here that attributing each encoder-block with a distinct attention-head ensures that there is a dedicated pathway to compute local attentions for every encoder-block. This allows us to parallelize the network and speed-up the decoding process when constructing the final summary.\looseness=-1\vspace{0.15cm}

\textbf{C.\hspace{0.2cm}The Decoder Layer: }Similar to the encoder, the decoder layer also consists of 3 parallelly stacked {decoder-blocks}. Each decoder-block contains an \textit{embedding layer} and a \textit{local-attention layer}. Parameters of the $i$-th encoder-block and~$i$-th decoder-block are shared. \looseness=-1
We construct a length-constrained summary~$s_b$ of the input document by processing each sentence in~$s_p$ sequentially. Depending on the remaining length-budget at each timestep, the final summary is constructed by soft-switching between a \textit{copy} and \textit{expand} operation. This process is guided by a sentence-level attention distribution~(Eq.~3) computed over~$s_p$. If the copy operation is selected, a sentence from~$s_p$ is copied into the final summary, whereas the expand operation replaces a sentence with similar content from the input document in~$s_b$. The original ordering of sentences is preserved.\vspace{0.15cm}\looseness=-1 
 

\textbf{The Copy Operation: }The probability of copying a sentence $s$ from the prototype summary that has not been included in the final summary~($s_b$) till timestep~$t$ into~$s_b$ is defined as follows: $P_c(s) = \vec{A}^{\,t}[s]$, where $\vec{A}^{\,t}$ represents a sentence-level attention distribution over~$s_p$ at timestep~$t$. Initialized as~$\vec{A^*}$ (Eq.~3), we update the attention distribution at each timestep after a copy or expand operation. If $s^*=argmax(P_c(s))$ represents the sentence copied into~$s_b$ at timestep~$t$, we update the attention distribution by zeroing out the probability of~$s^*$ in $\vec{A^t}$ and renormalizing the resulting distribution.\looseness=-1\vspace{0.15cm}

\textbf{The Expand Operation: } If the length of our prototype summary~($s_p$) is less than the length-budget~$b$, \texttt{MLS} can choose to expand a set of sentences from~$s_p$. For each sentence~$s\in s_p$, we define its \textit{expansion-set}~$E(s)$ as the sentence n-gram that is most similar to~$s$ in the input document. We determine the expansion-set~$E(s)$ of a sentence~$s$ by using beam-search over all $n$-grams in the input document that are yet to be included in the final summary. Our search objective being maximizing $score(E)=sim(s,E)\times overlap(s,E)$. The first term in $score(E)$ denotes the average pairwise cosine similarity between~$s$ and the sentences in~$E(s)$, whereas the second term denotes the fraction of tokens in~$s$ that appear in~$E(s)$. To minimize across-sentence repetitions in the summary, top 4 candidates identified from the search process are re-ranked~\cite{chen2018fast} based on the number of repeated word bigrams and trigrams if the expansion-set is included in the final summary. We obtained best performance by initializing $n$ with 3 and changing it to 2 at later iterations of the decoding process. If~$\vec{v}_i^{\,k}$ denotes the embedding-vector of the~$k$-th sentence in~$E(s)$ computed by the embedding-layer of the~$i$-th decoder-block, we define the probability of expanding a sentence~$s$ from the prototype summary to~$E(s)$ in the final summary as follows.\looseness=-1\vspace{-0.3cm}

\begin{gather}
\vec{C}_{i,k}^{\,e} = \frac{1}{r}\Sigma_{j=1}^r\hspace{0.05cm} dist_i(\vec{v_i}^{\,k},\hspace{0.05cm}\vec{Q_{i}}^{\,T}[j])\\
c_{i,k}^{\,e} = \frac{1}{n_i}\Sigma_{j=1}^{n_i} (\vec{C}_{i,k}^{\,e}[j])\\
\vec{A}^{\,e} = \frac{1}{m}\Sigma_{i=1}^{m}\hspace{0.05cm} (\vec{c}_i^{\,e}\cdot w_i) 
\end{gather}\vspace{-0.3cm}

In Eq.~4, $\vec{Q}_i$ denotes the semantic kernel shared between the~$i$-th encoder-block and decoder-block. We compute the probability of including the~$k^{th}$ sentence of~$E(s)$ into the final summary by computing its contribution~($c_{i,k}^e$) towards optimizing the qualitative property~$\Phi_i$ encoded by~$\vec{Q_i}$ first (Eq.~5). Repeating this process for all the sentences in~$E(s)$, followed by normalization provides us with the distribution $\vec{c}_i^{\,e} = (c_{i,1}^e, c_{i,2}^e, ...)$. Here, $\vec{c}_i^{\,e}$ represents the probability distribution over~$E(s)$. To obtain the expansion probability of a sentence in $E(s)$, we repeat this process for all 3 attention-heads and average them~(Eq.~6). The probability~$P_{e}(s)$ of expanding a sentence~$s$ from the prototype summary is obtained by averaging the expansion probability of all sentences in $E(s)$. Once a sentence~$s$ has been expanded into the final summary, we update the attention distribution by zeroing out the probability at~$s$ and renormalizing the resulting distribution.\looseness=-1\vspace{0.15cm} 

\textbf{Soft-Selection between Copy and Expansion: }We define the probability~$p_o(s)$ of selecting between the copy and expand operation for a sentence~$s$ in the prototype summary as follows.\looseness=-1\vspace{-0.4cm}

\begin{gather}
p_o(s) = \alpha \times P_e(s) + (1- \alpha)\times P_c(s)\\\vspace{0.05cm}
\alpha =\left\{
                \begin{array}{lll}
                0\hspace{3.1cm}if\hspace{0.1cm}b\leq len({s_b}^*)\\
                  \texttt{max}(P_e(s),P_c(s))\hspace{0.4cm} if\hspace{0.1cm}b > len({s_b}^*)\\
                \end{array}
              \right.
\end{gather}\vspace{-0.2cm}

In Eq.~8, ${s_b}^*$ denotes the partially constructed summary till timestep~$t$. If the length-budget~$b$ is smaller than the length of the prototype summary $s_p$, the probability of including a sentence from~$s_p$ into the final summary depends on the attention distribution~$\vec{A}^{\,t}$ over sentences in~$s_p$ that are not included in the final summary till timestep~$t$. In all other scenarios,~$\alpha$ acts as a soft-switch between copying or expanding a sentence in~$s_p$. A sentence can be expanded only if doing so does not exceed the length-budget. Once the probability of each sentence (and/or its expansion set) has been computed, the decoder attends to the position with the highest probability and copies/expands it into the final summary. Generation stops once $len({s_b}^*)$ reaches~$b$. We observed that the probability of expanding a sentence from the prototype summary (instead of copying it) increases with the allocated length-budget.\looseness=-1

\subsection{Training the Networks}\vspace{-0.1cm}
We trained PG-Network and the Pointer-Magnifier network separately on a NVIDIA Titan-XP GPU with a batch size of 16. We pretrained the PG-Network on the CNN-DailyMail dataset~\cite{nallapati2016abstractive} and then fine-tuned it on training samples of our experimental datasets. Using the evaluation script provided by~\cite{nallapati2016abstractive}, we obtained a training set of 287,226 pairs and validation set of 13,368 pairs for this dataset. All encoder-decoder weights were allowed to be updated during fine-tuning stage, following a L1-transfer~\cite{pan2009survey} of weights from the pretrained network. The external vocabulary used in both pretraining and fine-tuning stage consisted of 80K most frequent tokens in the training samples of the CNN-DailyMail dataset, our experimental dataset or both. Learning-rate and initial accumulator values were set to 0.15 and 0.1 respectively. We used Adagrad~\cite{duchi2011adaptive} to train the network. The encoder was fed a maximum 400 tokens and the decoder generated 100 tokens during pretraining. These values were increased to 500 and 200 respectively during fine-tuning. To prevent overfitting, we stopped training after 3000 iterations during the fine-tuning stage. With respect to the Pointer-Magnifier network, we learn the optimal values of~$w_i, 1\leq i\leq 3$ associated with each attention-head by grid-searching over the interval [-1,1] with the learning objective of maximizing ROUGE-1 score on the validation set. The optimal weights assigned to the attention-head corresponding to topic-coverage~($\phi_1$) and keyword-coverage~($\phi_2$) were positive, whereas information redundancy~($\phi_3$) was assigned a negative weight for both of our datasets.\looseness=-1

\begin{wraptable}{r}{0.5\linewidth}\vspace{-0.1cm}
 \centering
 
 \resizebox{\linewidth}{!}{%
 \renewcommand{\arraystretch}{0.8}
\begin{tabular}{lllcccc}
\toprule
\textbf{Index} & \textbf{Dataset}

& \multicolumn{1}{l}{\textbf{Size}}  & \multicolumn{1}{l}{\textbf{Max}}  & \multicolumn{1}{l}{\textbf{Median}} & \multicolumn{1}{l}{\textbf{Mean}}  \\ \midrule\midrule


\textbf{D1} & {MSR Narrative}     & 476                                                               & 130                                                         & 15    & 18.65                                \\
\midrule

\textbf{D2} & {Thinking Machines}  & 186                                                              & 82                                                         & 33  & 33.23  \\   
\midrule
\end{tabular}\vspace{-0.3cm}
}
  \caption{\textit{The minimum, maximum, median, and average number of sentences in datasets D1 and D2}\label{fig:datasetinfo}}\vspace{-0.2cm}
\end{wraptable}

\section{Experiments}\vspace{-0.1cm}
\label{s:experimental-design}
We seek to answer three key questions in our experiments. Given a length-constrained summary~$s_b$, (a) how similar is $s_b$ to a gold-standard summary?, (b) is it coherent and representative of the input document? and (c) how abstractive is~$s_b$? We answer the first two questions by evaluating the summaries generated by \texttt{MLS} over a range of success metrics on datasets belonging to two low-resource domains. We also conduct a user study to measure how representative are the summaries with respect to the input documents. A representative summary covers the main topics of the document. We answer the third question by computing the percentage of n-grams in~$s_b$ that do not appear in the input document and/or generated from the external vocabulary.\vspace{0.15cm}\looseness=-1

{\textbf{A.\hspace{0.2cm}Datasets}: }
We evaluate \texttt{MLS} on two publicly available datasets from two low-resource domains: the MSR-Narrative~\cite{ouyang2017crowd} (D1) dataset and the Thinking-Machines~\cite{thmch} (D2) dataset. The MSR-Narrative dataset contain personal stories shared by users on a social networking website. The Thinking-Machines dataset, on other hand, contains position papers on a popular scientific topic published in an educational website. Each document in both datasets is paired with a gold-standard summary. We randomly selected 25\% document-pairs to construct the training set and 10\% document-pairs to construct a validation set for both datasets. The rest comprised the test corpus. We present an overview of some of the important properties of both datasets in Table~\ref{fig:datasetinfo}.\looseness=-1\vspace{0.15cm}

\begin{table}[t]\resizebox{1\linewidth}{!}{
\centering\vspace{-0.4cm}
\renewcommand{\arraystretch}{0.9}
\begin{tabular}{|c|l|>{\columncolor[gray]{0.8}}cccc|>{\columncolor[gray]{0.8}}cccc|>{\columncolor[gray]{0.8}}cccc|>{\columncolor[gray]{0.8}}cccc|>{\columncolor[gray]{0.8}}cccc|}
\toprule
\multicolumn{1}{|c|}{\textbf{Dataset}}    & \textbf{Metric}           & \multicolumn{4}{c|}{\textbf{Budget = 1/32}} & \multicolumn{4}{c|}{\textbf{Budget = 1/16}} & \multicolumn{4}{c|}{\textbf{Budget = 1/8}} & \multicolumn{4}{c|}{\textbf{Budget = 1/4}}           & \multicolumn{4}{c|}{\textbf{Budget = 1/2}}            \\
\multicolumn{1}{|l|}{}                    &                           & \textbf{MLS}              & \textbf{A1}   & \textbf{A2}   &  \textbf{A3} & \textbf{MLS}             & \textbf{A1}      & \textbf{A2} &  \textbf{A3} & \textbf{MLS}      & \textbf{A1}      & \textbf{A2} & \textbf{A3} & \textbf{MLS}            & \textbf{A1}    & \textbf{A2}   & \textbf{A3}    & \textbf{MLS}          & \textbf{A1}     & \textbf{A2}     & \textbf{A3}  \\ \midrule

\multirow{4}{*}{\textbf{D1}} & \textcolor{black}{\textbf{ROUGE-1}}                   & \textbf{45.99}   & 23.44        & 37.46      & 41.65 & \textbf{45.99}  & 30.50    & 37.68     & 43.07 & \textbf{45.99}  & 31.27   & 38.05 & 43.50   & \textbf{46.11} & 41.86  & 43.95 &  44.10 & \textbf{45.67} & {40.67} & 41.13    & {45.50}      \\
                                        & \textcolor{black}{\textbf{ROUGE-2}}  & \textbf{35.97}   & 14.79  & 22.59  & 30.65     & \textbf{35.97}  & 20.77  & 25.50 & 30.65     & \textbf{35.98}  & 22.95  & 29.14  & 33.50  & \textbf{35.60}  & 27.57  & 32.36  &  34.50    & \textbf{36.70}  & 29.38 & 31.02   &  35.02   \\
                                        & \textcolor{black}{\textbf{ROUGE-L}}                   & \textbf{40.89}  & 21.35        & 32.38      & 37.65 & \textbf{42.50}     & 27.9       & 33.07     &  38.92 & {43.01}     & 36.25      & 37.62   & \textbf{43.50} & \textbf{42.83} & 38.83     & 40.95    &  41.07     & 40.18           & 39.60           & {40.74} & \textbf{41.50}\\
                                        & \textcolor{black}{\textbf{METEOR}}                    & \textbf{47.12}   & 18.91   & 24.22    & 45.51   & \textbf{47.12}  & 13.07   & 25.02  &  45.60 & \textbf{46.50}   & 20.89   & 30.86  & 43.88   & \textbf{46.61} & 27.26   & 33.05 &  44.65  & \textbf{45.71} & {27.84}  &   32.95   &  {45.39}   \\
                                        \midrule\midrule 

\multirow{4}{*}{\textbf{D2}} & \textcolor{black}{\textbf{ROUGE-1}}                   & \textbf{40.25}   & 16.20        & 21.06  & 35.60    & \textbf{40.0}  & 17.08     & 22.0  &  36.0    & \textbf{40.25}  & 22.59   & 28.10  &  39.72   & \textbf{41.01} & 23.55  & 27.83  &  38.50  & \textbf{44.36} & 29.53 &  32.75  & 44.06 \\
                                        & \textcolor{black}{\textbf{ROUGE-2}}     & \textbf{33.25}   & 11.25  & 17.22    &   26.50  & \textbf{34.50}  & 12.0  & 16.75   &  30.05  & \textbf{35.67}  & 14.60  &  19.01  &  31.80  & \textbf{36.0}  & 17.90  & 20.06  &  {31.0}    & \textbf{38.70}  & 20.67 & 23.46  &  36.44   \\
                                        & \textcolor{black}{\textbf{ROUGE-L}}                   & \textbf{37.17}  & 14.50  & 19.06  &  33.67   &
                                        \textbf{37.0}  & 15.60   & 20.55  &  35.70    & \textbf{37.05}  & 21.65  & 20.26  & 34.33  & \textbf{37.96}  & 21.87  & 22.60  &  32.77   & \textbf{41.50 } & 26.04 &  27.17 &  39.75\\
                                        & \textcolor{black}{\textbf{METEOR}}                    & \textbf{40.22}   & 12.68   & 24.33   &    35.05   & \textbf{44.82}  & 15.17  & 23.22  &  42.90  & \textbf{44.82}  & 11.96 & 30.79  &  42.0    & \textbf{42.88} & 24.20 & 21.83  & 38.05 & {44.79} & 28.08 & 25.82  & \textbf{45.70}   \\
                                        \midrule
\end{tabular}}\vspace{-0.2cm}

\begin{small}
\caption{\textit{ROUGE and METEOR scores of the budgeted summaries constructed by} \texttt{MLS}~\textit{(highlighted column) and the baseline methods for the MSR-Narrative (D1) and Thinking Machines (D2) dataset\looseness=-1}
\label{table:cnn_msr_result}
}
\end{small}
\vspace{-0.4cm}

\end{table}

{\textbf{B.\hspace{0.2cm} Metrics}: }We compare the summaries constructed by \texttt{MLS} against gold-standard summaries using METEOR~\cite{banerjee2005meteor} and ROUGE~\cite{lin2004rouge} scores\footnote{We used py-rouge~\cite{pyrouge} and the NLTK library to compute the ROUGE and METEOR score respectively.}. The average $F_1$ score of ROUGE-1, ROUGE-2 and ROUGE-L metrics obtained for both datasets are shown in Table~\ref{table:cnn_msr_result}. 
To measure the \textit{representativeness} of a summary, we compute the average KL-divergence score between the top-3 topic vectors of a summary and its input document. Following~\cite{srinivasan2018corpus}, we measure the coherence of a summary by computing the average cosine similarity between consecutive sentences. We report the absolute difference between the coherence score computed for a summary and its input document in Table~\ref{table:cnn_msr_results_coherence}. We also report the KL-divergence score between sentiment vectors of a summary and the input document to check for potential biases in its polarity distribution. {We used a publicly available library~\cite{hutto2014vader} to derive the sentiment vectors.}. Note that, lower values of $\Delta Coherence$ and KL-divergence score are desirable for a high-quality summary.\looseness=-1\vspace{0.15cm}

\textbf{C.\hspace{0.2cm} Baselines: }We compare \texttt{MLS} against three baseline methods. Two of them follow a sampling based approach, while our final baseline method employs a convolutional network to construct length budgeted summaries. Our first baseline~(\textbf{A1}) follows a systematic sampling based approach to construct length-controlled summaries. Initialized with a randomly selected sentence from the first~$k$-1 sentences of the input document, it constructs the final summary by including the~$k$-th sentence from the last sampled position. We set $k$ = 3 in all of our experiments for both datasets. Sampling terminates when the budget limit is exceeded or the end of document is reached. Our second baseline method~(\textbf{A2}) follows a weighted graph-based sampling strategy to construct budgeted summaries. It represents each sentence in the input document as a node in an undirected, complete, weighted graph. The weight assigned to an edge in this graph is equal to the pairwise cosine similarity between the connecting nodes. To construct the budgeted summary, we sample the top-$K$ nodes of this graph using a weighted PageRank algorithm~\cite{mihalcea2004textrank}. Sampling stops when the budget is reached. Our third and final baseline method~(\textbf{A3}) is a convolutional approach proposed in~\cite{liu2018controlling}. It is a sequence-to-sequence network with Gated Linear Units~\cite{dauphin2017language} that takes the desired length of a summary as an additional input to the initial state of the decoder network. Similar to our training protocol, we pretrain this network on the CNN-DailyMail dataset first and fine-tune it on the training samples from both of our experimental datasets. We allowed all weights to be updated during the fine-tuning phase.\looseness=-1

\subsection{Results and Discussion}
\label{s:expr-results}
We report the performance of all competing methods at five length-budgets. We specify the length-budget to construct a summary as a product of the number of tokens in the input document and a compression-budget $c \in \{\frac{1}{32}, \frac{1}{16}, \frac{1}{8}, \frac{1}{4}, \frac{1}{2} \}$. Results from our experiments are presented in Tables~\ref{table:cnn_msr_result} and~\ref{table:cnn_msr_results_coherence}. The best performance achieved for each metric is boldfaced. We highlight some of our key findings below.

\subsubsection{Qualitative Evaluation at five Compression Budgets}
In general, the abstractive methods (\texttt{MLS} and A3) outperform sampling-based approaches~(see Table~\ref{table:cnn_msr_result}) on both datasets. \texttt{MLS} performs consistently well on all budgets, although performance is relatively better on smaller budgets. We obtain an absolute improvement of 4.34\% and 4.65\% in ROUGE-1 score \& 1.61\% and 5.17\% in METEOR score over the convolutional baseline~(A3) for datasets D1 and D2 at compression budget = $\frac{1}{32}$. At higher budgets, our performance was comparable with A3. In terms of coherence, \texttt{MLS} performs comparably or better than A3~(see Table~\ref{table:cnn_msr_results_coherence}). Smaller $\Delta Coherence$ score than A1 and A2 suggests that \texttt{MLS} generated more coherent summaries than these two baseline methods. Small KL-divergence between the topic distribution of a budgeted summary and input document shows that \texttt{MLS} generated summaries are {representative} of the document for both datasets. In fact, topic-coverage in summaries generated by \texttt{MLS} is at least 75\% better than the convolutional baseline~(A3)~\cite{liu2018controlling}, although performance becomes comparable at larger budgets as more sentences from the prototype summary are expanded to make the final summary. \texttt{MLS} outperfoms A1 and A2 in terms of staying true to the sentiment distribution of the input document. This can be seen from the small KL-divergence scores obtained for the sentiment distribution achieved by \texttt{MLS} in Table~\ref{table:cnn_msr_results_coherence}.\looseness=-1 

\begin{wrapfigure}{r}{0.65\linewidth}
\vspace{-0.4cm}

\begin{center}
    \includegraphics[width=\linewidth]{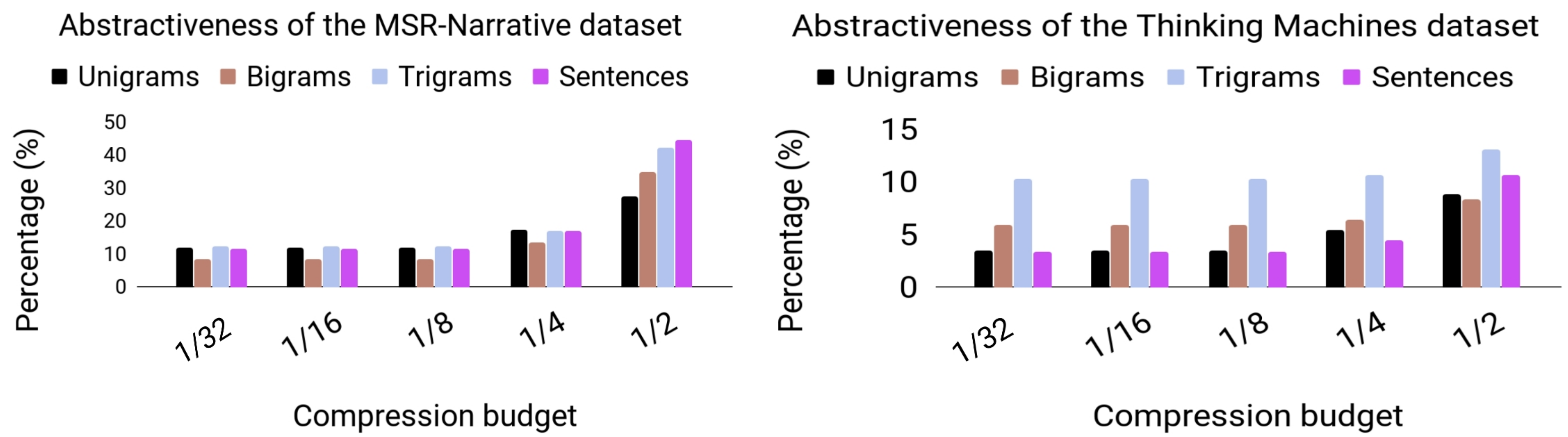}
\end{center}\vspace{-0.2cm}
\begin{small}
\caption{\textit{Abstractiveness of} \texttt{MLS} \textit{generated summaries}}\vspace{-0.3cm}
\label{fig:abstractiveness}
\end{small}

\end{wrapfigure}

\texttt{MLS} generated summaries were more abstractive at higher budgets~(Fig.~\ref{fig:abstractiveness}). At compression budget = $\frac{1}{2}$, 27.35\% tokens in the summaries constructed for dataset D1 and 8.75\% tokens for dataset D2 were contributed by the external vocabulary.\looseness=-1


\begin{table}[t]\resizebox{1\linewidth}{!}{
\vspace{-0.4cm}
\centering
\renewcommand{\arraystretch}{0.9}
\begin{tabular}{|c|l|>{\columncolor[gray]{0.8}}cccc|>{\columncolor[gray]{0.8}}cccc|>{\columncolor[gray]{0.8}}cccc|>{\columncolor[gray]{0.8}}cccc|>{\columncolor[gray]{0.8}}cccc|}
\toprule
\multicolumn{1}{|c|}{\textbf{Dataset}}    & \textbf{Metric}           & \multicolumn{4}{c|}{\textbf{Budget = 1/32}} & \multicolumn{4}{c|}{\textbf{Budget = 1/16}} & \multicolumn{4}{c|}{\textbf{Budget = 1/8}} & \multicolumn{4}{c|}{\textbf{Budget = 1/4}}           & \multicolumn{4}{c|}{\textbf{Budget = 1/2}}            \\
\multicolumn{1}{|l|}{}                    &                           & \textbf{MLS}     & \textbf{A1}   & \textbf{A2} & \textbf{A3}   & \textbf{MLS}    & \textbf{A1}      & \textbf{A2}  &   \textbf{A3}  & \textbf{MLS}      & \textbf{A1}      & \textbf{A2} & \textbf{A3} & \textbf{MLS}  & \textbf{A1}  & \textbf{A2} & \textbf{A3}  & \textbf{MLS}  & \textbf{A1}     & \textbf{A2}  & \textbf{A3}       \\ \midrule

\multirow{3}{*}{\textbf{D1}} & \textcolor{black}{\textbf{Topic}} & \textbf{0.12}    & 0.28         & 0.29   &  0.21    & \textbf{0.12}   & 0.27     & 0.27   &  0.20    & \textbf{0.12}   & 0.26    & 0.23  &  0.15   & \textbf{0.13}  & 0.21   & 0.19  &  0.18  & \textbf{0.13}  & 0.21           & 0.21  & 0.18    \\
                                        & \textcolor{black}{\textbf{Sentiment}} & \textbf{0.09}    & 0.22    & 0.19  & 0.11      & \textbf{0.09}   & 0.23       & 0.15  &  0.13  & \textbf{0.09}   & 0.19  & 0.15  & 0.12   & \textbf{0.1}   & 0.14     & 0.12  &  \textbf{0.1}   & 0.16    & \textbf{0.07}  & 0.17  &  0.13   \\
                                        & \textcolor{black}{\textbf{$\Delta$Coherence}} & \textbf{0.08}    & 0.3          & 0.20    &  0.11   & \textbf{0.08}   & 0.26  & 0.18 & 0.09  & {0.08}   & 0.21  & 0.11 & \textbf{0.07}     & \textbf{0.09}  & 0.13    & 0.10  &  0.12  & 0.1 & \textbf{0.06}  & 0.09 & 0.1 \\\midrule\midrule

\multirow{3}{*}{\textbf{D2}} & \textcolor{black}{\textbf{Topic}} & \textbf{0.05}  & 0.27  & 0.24 & 0.15 & \textbf{0.05}   & 0.27     &  0.25    & 0.16 & \textbf{0.05}   & 0.17     & 0.2 & 0.12 & \textbf{0.05}  & 0.08  & 0.08  & 0.11   & {0.03}   & {0.03}  & \textbf{0.02} & 0.10   \\
                                        & \textcolor{black}{\textbf{Sentiment}} & \textbf{0.03}    & 0.24        & 0.16   & 0.10 & \textbf{0.03}  & 0.21    & 0.13  &  0.07    & \textbf{0.03}   & 0.12  & 0.15 & 0.04  & \textbf{0.03}   & 0.06  & 0.08  & 0.05  & 0.04    & \textbf{0.02}     & 0.03    & 0.03       \\
                                        & \textcolor{black}{\textbf{$\Delta$Coherence}} & \textbf{0.03}    & 0.27     & 0.20  & 0.05    & \textbf{0.03}   & 0.18      & 0.12 & 0.10  & \textbf{0.03}   & 0.09    & 0.09 & 0.05 & \textbf{0.03}  & 0.05  & 0.05  & 0.06 & 0.04   & \textbf{0.03}  & \textbf{0.03} & 0.04  \\\midrule

\end{tabular}}
\begin{small}

\caption{\textit{Coherence and completeness of the budgeted summaries constructed by} \texttt{MLS}~\textit{(highlighted column) and the baseline methods for MSR-Narrative~(D1) and Thinking Machines~(D2) dataset\looseness=-1}}\vspace{-0.3cm}
\label{table:cnn_msr_results_coherence}
\end{small}
\end{table}
\subsubsection{Ablative Analysis}
\begin{wrapfigure}{r}{0.65\linewidth}

\begin{center}
    \includegraphics[width=\linewidth]{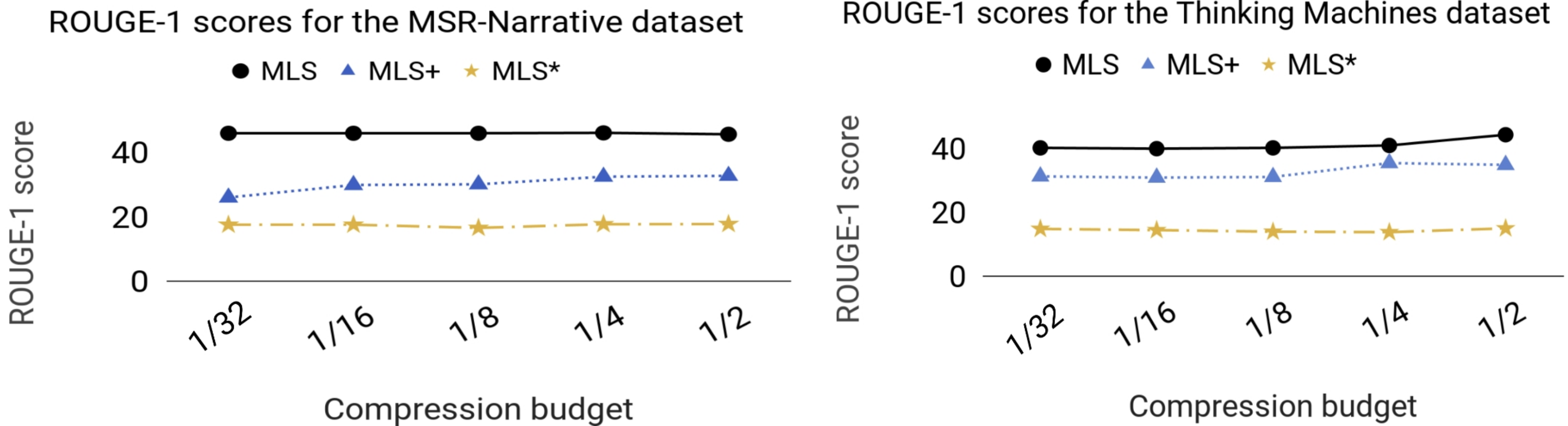}
\end{center}
\begin{small}\vspace{-0.2cm}
\caption{\textit{ROUGE-1 score of} \texttt{MLS} \textit{and the ablative baselines MLS+ and MLS* on datasets D1 and D2}}\vspace{-0.1cm}
\label{fig:pretraining}
\end{small}

\end{wrapfigure}

To investigate the effects of pretraining on end-to-end results, we compare the ROUGE-1 score of summaries constructed by \texttt{MLS} against an ablative baseline~{MLS}*. It is identical to \texttt{MLS} except that the PG-Network was not pretrained. In our second experiment, we compare \texttt{MLS} against MLS+, an ablative baseline that constructs the prototype summary following a greedy heuristics~\cite{otterbacher2006news} instead of the PG-network. \texttt{MLS} outperforms both baselines (Fig.~\ref{fig:pretraining}) on both datasets, thereby establishing that using PG-Network in our framework and pretraining it on the CNN-DailyMail dataset improved the quality of our final summaries. Finally, to investigate the effects of the semantic kernels introduced in the Pointer-Magnifier network, we iteratively replaced each of the three semantic kernels~(Section 2.2) with a randomized kernel by shuffling its rows and columns. 

We observed an absolute decrease of up to 4.30\% in ROUGE-1 score and 3.75\% in METEOR score for~$\Vec{Q_3}$, with bigger impacts in performance at higher length-budgets. Replacing~$\Vec{Q_2}$ with a randomized kernel, on other hand, decreased the average $\Delta Coherence$ score by approximately 45\% for dataset D1 and 30\% for D2 for summaries constructed at compression budget = $\frac{1}{2}$, i.e. half-length of the input document.\looseness=-1

\subsubsection{Human Evaluation of Length-Controlled Summaries}
We conducted a study to evaluate the completeness of the summaries constructed by \texttt{MLS}. More specifically, we considered a scenario where the user needs to complete a fact checking task. We chose three documents from both datasets randomly and asked each participant to verify the presence of some key facts of the document in the summaries constructed by \texttt{MLS} and/or a baseline method. Each participant was instructed to complete the task solely based on the content of the summary and not depending on any previous knowledge. For example, the question  ``\textit{Does the story tell us why the narrator was fired?}" was paired with the following summary-- ``\textit{I tried to return a lost wallet to a customer who accused me of stealing it and then grabbed my hair. We got in a physical fight and I was fired from my job}". The participants had to chose between `Yes', `No", and ``More information required". If a participant selected the third option, a longer summary was shown with the same question. The task was terminated otherwise. In addition to \texttt{MLS}, A2 (the stronger extractive baseline in our experimental setup) and A3, we add two extreme settings: (a) the full-content setting in which the original document was shown, and (b) the no-content setting where no textual content (other than the question itself) was shown to a participant. The full-content setting ensured that the question could indeed be answered from the article, whereas the no-content setting ensured whether the questions contained any hint about the answer. 

\begin{wraptable}{r}{0.55\linewidth}
 \centering
 
 \resizebox{\linewidth}{!}{%
 \renewcommand{\arraystretch}{0.8}
\begin{tabular}{ll>{\columncolor[gray]{0.8}}ccccc}

\toprule
\textbf{Index} & \textbf{Dataset} & \textbf{MLS} & \textbf{A2}           & \textbf{A3} & \multicolumn{1}{c}{\textbf{NC}} & \multicolumn{1}{c}{\textbf{FC}} \\ \midrule \midrule
\multirow{2}{*}{\textbf{D1}} & \textbf{Accuracy}  & \textbf{0.88} & 0.55          & 0.55          & 0.0                            & \textbf{0.88}                    \\ 
                         & \textbf{Duration (s)} & \textbf{36.7} & 43.69         & 69.08         & 12.0                           & 75.6                             \\   \midrule 
\multirow{2}{*}{\textbf{D2}}  & \textbf{Accuracy} & 0.55          & 0.44          & \textbf{0.66} & 0.0                            & \textbf{0.88}                    \\
                         & \textbf{Duration (s)} & 70.24         & \textbf{68.9} & 96.47         & 20.95                          & 132.86      \\ \midrule                 
\end{tabular}
}\vspace{-0.25cm}

\caption{\textit{Mean accuracy and completion time using} \texttt{MLS}, \textit{A2, A3, No-content (NC) and Full-content (FC) settings\looseness=-1}\label{table:userstudy_results}}\label{t:user_study_results}\vspace{-0.25cm}
\end{wraptable}

The task started by showing each participant a summary generated at compression budget = $1/32$. If they opted for more information to be shown, we provided a summary generated by the same method by doubling the compression budget each time until the user responded with a `Yes' or `No' or we reached the budget of $1/2$. The key intuition here is that if users are given a complete and representative summary, they should be able to answer the questions accurately, as a good summarization model would pick up the key concepts of the document even at shorter length-budgets, without requiring for it to be expanded further. With this in mind, we recorded task completion time and user response for each treatment. All budgeted summaries were constructed beforehand. We invited 22 graduate students to participate in the study. Each participant was shown summaries generated by at most two different methods in random order. No information on the method used was revealed to a participant at any stage. To prevent information retention, each participant was shown a summary generated from the same document only once. Using a balanced, incomplete block design~\cite{aschbacher1971collineation}, each of the 10 settings (5 methods $\times$ 2 datasets) was assigned to 3 subjects. The average accuracy and task completion time recorded for each treatment is shown in Table~\ref{t:user_study_results}. The accuracy of the no-content setting is 0 for both datasets, indicating that the questions did not contain any hint to the correct answer, whereas the full-content setting shows that overall the questions could have been answered from the original documents. 
When using summaries generated by \texttt{MLS}, the participants responded as accurately as the Full-content setting on dataset D1, while being more than two times faster, outperforming A2 and A3. 
For dataset D2, participants were more accurate using summaries constructed by \texttt{MLS} than A2. MLS performed better than A3 on one document, comparable on one and worse on one document, with an average accuracy of 0.55.\looseness=-1 

\section{Conclusion}\vspace{-0.2cm}
We have proposed \texttt{MLS}, a supervised approach to construct abstractive summaries at controllable lengths. Following an extract-then-compress paradigm, we develop the Pointer-Magnifier network -- a length-aware, encoder-decoder network that constructs length-constrained summaries by shortening or expanding a prototype summary inferred from the document. The key enabler of this network is an array of semantic kernels with clearly defined human-interpretable syntactic/semantic roles in constructing the summary given a budget-length. We train our network on limited training samples from two cross-domain datasets. Experiments show that the summaries constructed by \texttt{MLS} are coherent and reflectively capture the main concepts of the document. Our human evaluation study also suggest the same. In the future, we would like to extend our work to construct task-driven summaries for interactive question answering tasks. Personalizing a summary based on user's past interaction model is another exciting direction of future work.\looseness=-1        
\section{Acknowledgement}\vspace{-0.2cm}

We would like to thank Professors Micha Elsner, Joel Bloch, Marie-Catherine de Marneffe, and Michael White for valuable discussions and comments. We would also like to thank the reviewers and folks who participated in our study for sharing critical feedback that helped improve our work. 

\begin{small}
\bibliographystyle{coling}
\bibliography{mybib}

\begin{thebibliography}{}

\bibitem[\protect\citename{Aschbacher}1971]{aschbacher1971collineation}
Michael Aschbacher.
\newblock 1971.
\newblock On collineation groups of symmetric block designs.
\newblock {\em Journal of Combinatorial Theory, Series A}.

\bibitem[\protect\citename{Baan \bgroup et al.\egroup
  }2019]{baan2019transformer}
Joris Baan, Maartje ter Hoeve, Marlies van~der Wees, Anne Schuth, and Maarten
  de~Rijke.
\newblock 2019.
\newblock Do transformer attention heads provide transparency in abstractive
  summarization?
\newblock {\em arXiv preprint arXiv:1907.00570}.

\bibitem[\protect\citename{Banerjee and Lavie}2005]{banerjee2005meteor}
Satanjeev Banerjee and Alon Lavie.
\newblock 2005.
\newblock Meteor: An automatic metric for mt evaluation with improved
  correlation with human judgments.
\newblock In {\em Proceedings of the acl workshop on intrinsic and extrinsic
  evaluation measures for machine translation and/or summarization}.

\bibitem[\protect\citename{Benjamin~Heinzerling}2020]{pyrouge}
Anders~Johannsen Benjamin~Heinzerling.
\newblock 2020.
\newblock A python wrapper for the rouge summarization evaluation package.
\newblock \url{https://pypi.org/project/pyrouge/}.

\bibitem[\protect\citename{Blei \bgroup et al.\egroup }2003]{blei2003latent}
David~M Blei, Andrew~Y Ng, and Michael~I Jordan.
\newblock 2003.
\newblock Latent dirichlet allocation.
\newblock {\em JMLR}.

\bibitem[\protect\citename{Brockman}2018]{thmch}
John Brockman.
\newblock 2018.
\newblock Thinking machines.
\newblock
  \url{https://www.edge.org/annual-question/what-do-you-think-about-machines-that-think}.

\bibitem[\protect\citename{Buyukkokten \bgroup et al.\egroup
  }2001]{buyukkokten2001seeing}
Orkut Buyukkokten, Hector Garcia-Molina, and Andreas Paepcke.
\newblock 2001.
\newblock Seeing the whole in parts: text summarization for web browsing on
  handheld devices.
\newblock In {\em The Web Conference}.

\bibitem[\protect\citename{Campana and Tombros}2009]{campana2009incremental}
Marco Campana and Anastasios Tombros.
\newblock 2009.
\newblock Incremental personalised summarisation with novelty detection.
\newblock In {\em FQAS}.

\bibitem[\protect\citename{Chen and Bansal}2018]{chen2018fast}
Yen-Chun Chen and Mohit Bansal.
\newblock 2018.
\newblock Fast abstractive summarization with reinforce-selected sentence
  rewriting.
\newblock {\em arXiv preprint:1805.11080}.

\bibitem[\protect\citename{Dauphin \bgroup et al.\egroup
  }2017]{dauphin2017language}
Yann~N Dauphin, Angela Fan, Michael Auli, and David Grangier.
\newblock 2017.
\newblock Language modeling with gated convolutional networks.
\newblock In {\em ICML}.

\bibitem[\protect\citename{Doshi-Velez and Kim}2017]{doshi2017towards}
Finale Doshi-Velez and Been Kim.
\newblock 2017.
\newblock Towards a rigorous science of interpretable machine learning.
\newblock {\em arXiv preprint arXiv:1702.08608}.

\bibitem[\protect\citename{Duchi \bgroup et al.\egroup
  }2011]{duchi2011adaptive}
John Duchi, Elad Hazan, and Yoram Singer.
\newblock 2011.
\newblock Adaptive subgradient methods for online learning and stochastic
  optimization.
\newblock {\em JMLR}.

\bibitem[\protect\citename{Fan \bgroup et al.\egroup
  }2017]{fan2017controllable}
Angela Fan, David Grangier, and Michael Auli.
\newblock 2017.
\newblock Controllable abstractive summarization.
\newblock {\em arXiv preprint:1711.05217}.

\bibitem[\protect\citename{Gehring \bgroup et al.\egroup
  }2017]{gehring2017convolutional}
Jonas Gehring, Michael Auli, David Grangier, Denis Yarats, and Yann~N Dauphin.
\newblock 2017.
\newblock Convolutional sequence to sequence learning.
\newblock In {\em ICML}. JMLR. org.

\bibitem[\protect\citename{Hermann \bgroup et al.\egroup
  }2015]{hermann2015teaching}
Karl~Moritz Hermann, Tomas Kocisky, Edward Grefenstette, Lasse Espeholt, Will
  Kay, Mustafa Suleyman, and Phil Blunsom.
\newblock 2015.
\newblock Teaching machines to read and comprehend.
\newblock In {\em Advances in neural information processing systems}, pages
  1693--1701.

\bibitem[\protect\citename{Hutto and Gilbert}2014]{hutto2014vader}
Clayton~J Hutto and Eric Gilbert.
\newblock 2014.
\newblock Vader: A parsimonious rule-based model for sentiment analysis of
  social media text.
\newblock In {\em ICWSM}.

\bibitem[\protect\citename{ITGP}2017]{team2017eu}
ITGP.
\newblock 2017.
\newblock {\em EU General Data Protection Regulation (GDPR).}
\newblock IT Governance Limited.

\bibitem[\protect\citename{Jain and Wallace}2019]{jain2019attention}
Sarthak Jain and Byron~C Wallace.
\newblock 2019.
\newblock Attention is not explanation.
\newblock In {\em Proceedings of the 2019 Conference of the North American
  Chapter of the Association for Computational Linguistics: Human Language
  Technologies, Volume 1 (Long and Short Papers)}, pages 3543--3556.

\bibitem[\protect\citename{Keymanesh \bgroup et al.\egroup
  }2020]{keymaneshtoward}
Moniba Keymanesh, Micha Elsner, and Srinivasan Parthasarathy.
\newblock 2020.
\newblock Toward domain-guided controllable summarization of privacy policies.
\newblock {\em Natural Legal Language Processing Workshop at KDD}.

\bibitem[\protect\citename{Kikuchi \bgroup et al.\egroup
  }2016]{kikuchi2016controlling}
Yuta Kikuchi, Graham Neubig, Ryohei Sasano, Hiroya Takamura, and Manabu
  Okumura.
\newblock 2016.
\newblock Controlling output length in neural encoder-decoders.
\newblock {\em arXiv preprint:1609.09552}.

\bibitem[\protect\citename{Le and Mikolov}2014]{le2014distributed}
Quoc Le and Tomas Mikolov.
\newblock 2014.
\newblock Distributed representations of sentences and documents.
\newblock In {\em ICML}.

\bibitem[\protect\citename{Lin}2004]{lin2004rouge}
Chin-Yew Lin.
\newblock 2004.
\newblock Rouge: A package for automatic evaluation of summaries.
\newblock {\em Text Summarization Branches Out}.

\bibitem[\protect\citename{Liu \bgroup et al.\egroup }2018a]{liu2018toward}
Fei Liu, Jeffrey Flanigan, Sam Thomson, Norman Sadeh, and Noah~A Smith.
\newblock 2018a.
\newblock Toward abstractive summarization using semantic representations.
\newblock {\em arXiv preprint:1805.10399}.

\bibitem[\protect\citename{Liu \bgroup et al.\egroup
  }2018b]{liu2018controlling}
Yizhu Liu, Zhiyi Luo, and Kenny Zhu.
\newblock 2018b.
\newblock Controlling length in abstractive summarization using a convolutional
  neural network.
\newblock In {\em EMNLP}.

\bibitem[\protect\citename{Liu \bgroup et al.\egroup }2019]{liu2019automatic}
Chunyi Liu, Peng Wang, Jiang Xu, Zang Li, and Jieping Ye.
\newblock 2019.
\newblock Automatic dialogue summary generation for customer service.
\newblock In {\em SIGKDD}.

\bibitem[\protect\citename{Magooda and Litman}2020]{magooda2020abstractive}
Ahmed Magooda and Diane Litman.
\newblock 2020.
\newblock Abstractive summarization for low resource data using domain transfer
  and data synthesis.
\newblock {\em arXiv preprint arXiv:2002.03407}.

\bibitem[\protect\citename{Mihalcea and Tarau}2004]{mihalcea2004textrank}
Rada Mihalcea and Paul Tarau.
\newblock 2004.
\newblock Textrank: Bringing order into text.
\newblock In {\em EMNLP}.

\bibitem[\protect\citename{Moradi and Ghadiri}2018]{moradi2018different}
Milad Moradi and Nasser Ghadiri.
\newblock 2018.
\newblock Different approaches for identifying important concepts in
  probabilistic biomedical text summarization.
\newblock {\em Artificial intelligence in medicine}, 84:101--116.

\bibitem[\protect\citename{Nallapati \bgroup et al.\egroup
  }2016]{nallapati2016abstractive}
Ramesh Nallapati, Bowen Zhou, Caglar Gulcehre, Bing Xiang, et~al.
\newblock 2016.
\newblock Abstractive text summarization using sequence-to-sequence rnns and
  beyond.
\newblock {\em arXiv preprint arXiv:1602.06023}.

\bibitem[\protect\citename{Otterbacher \bgroup et al.\egroup
  }2006]{otterbacher2006news}
Jahna Otterbacher, Dragomir Radev, and Omer Kareem.
\newblock 2006.
\newblock News to go: hierarchical text summarization for mobile devices.
\newblock In {\em SIGIR}.

\bibitem[\protect\citename{Ouyang \bgroup et al.\egroup }2017]{ouyang2017crowd}
Jessica Ouyang, Serina Chang, and Kathy McKeown.
\newblock 2017.
\newblock Crowd-sourced iterative annotation for narrative summarization
  corpora.
\newblock In {\em EACL}.

\bibitem[\protect\citename{Pan and Yang}2009]{pan2009survey}
Sinno~Jialin Pan and Qiang Yang.
\newblock 2009.
\newblock A survey on transfer learning.
\newblock {\em TKDE}.

\bibitem[\protect\citename{Parida and
  Motlicek}2019]{parida-motlicek-2019-abstract}
Shantipriya Parida and Petr Motlicek.
\newblock 2019.
\newblock Abstract text summarization: A low resource challenge.
\newblock In {\em EMNLP}. Association for Computational Linguistics, November.

\bibitem[\protect\citename{Paulus \bgroup et al.\egroup }2018]{paulus2018deep}
Romain Paulus, Caiming Xiong, and Richard Socher.
\newblock 2018.
\newblock A deep reinforced model for abstractive summarization.
\newblock In {\em International Conference on Learning Representations}.

\bibitem[\protect\citename{Rose \bgroup et al.\egroup }2010]{rose2010automatic}
Stuart Rose, Dave Engel, Nick Cramer, and Wendy Cowley.
\newblock 2010.
\newblock Automatic keyword extraction from individual documents.
\newblock {\em Text mining: applications and theory}.

\bibitem[\protect\citename{Rush \bgroup et al.\egroup }2015]{rush2015neural}
Alexander~M Rush, Sumit Chopra, and Jason Weston.
\newblock 2015.
\newblock A neural attention model for abstractive sentence summarization.
\newblock {\em arXiv preprint:1509.00685}.

\bibitem[\protect\citename{Saito \bgroup et al.\egroup }2020]{saito2020length}
Itsumi Saito, Kyosuke Nishida, Kosuke Nishida, Atsushi Otsuka, Hisako Asano,
  Junji Tomita, Hiroyuki Shindo, and Yuji Matsumoto.
\newblock 2020.
\newblock Length-controllable abstractive summarization by guiding with summary
  prototype.
\newblock {\em arXiv preprint:2001.07331}.

\bibitem[\protect\citename{Sarkhel and Nandi}2019]{sarkhel2019visual}
Ritesh Sarkhel and Arnab Nandi.
\newblock 2019.
\newblock Visual segmentation for information extraction from heterogeneous
  visually rich documents.
\newblock In {\em Proceedings of the 2019 International Conference on
  Management of Data}, pages 247--262.

\bibitem[\protect\citename{Sarkhel \bgroup et al.\egroup
  }2018]{sarkhel2018nurses}
Ritesh Sarkhel, Jacob~J Socha, Austin Mount-Campbell, Susan Moffatt-Bruce,
  Simon Fernandez, Kashvi Patel, Arnab Nandi, and Emily~S Patterson.
\newblock 2018.
\newblock How nurses identify hospitalized patients on their personal notes:
  Findings from analyzing ‘brains’ headers with multiple raters.
\newblock In {\em Proceedings of the International Symposium on Human Factors
  and Ergonomics in Health Care}, volume~7, pages 205--209. SAGE Publications
  Sage CA: Los Angeles, CA.

\bibitem[\protect\citename{See \bgroup et al.\egroup }2017]{see2017get}
Abigail See, Peter~J Liu, and Christopher~D Manning.
\newblock 2017.
\newblock Get to the point: Summarization with pointer-generator networks.
\newblock {\em arXiv preprint:1704.04368}.

\bibitem[\protect\citename{Song \bgroup et al.\egroup
  }2020a]{joint-parsing-summarization:2020}
Kaiqiang Song, Logan Lebanoff, Qipeng Guo, Xipeng Qiu, Xiangyang Xue, Chen Li,
  Dong Yu, and Fei Liu.
\newblock 2020a.
\newblock Joint parsing and generation for abstractive summarization.
\newblock In {\em Proceedings of the AAAI Conference on Artificial
  Intelligence}.

\bibitem[\protect\citename{Song \bgroup et al.\egroup
  }2020b]{control-over-copying:2020}
Kaiqiang Song, Bingqing Wang, Zhe Feng, Liu Ren, and Fei Liu.
\newblock 2020b.
\newblock Controlling the amount of verbatim copying in abstractive
  summarization.
\newblock In {\em Proceedings of the AAAI Conference on Artificial
  Intelligence}.

\bibitem[\protect\citename{Srinivasan \bgroup et al.\egroup
  }2018]{srinivasan2018corpus}
Balaji~Vasan Srinivasan, Pranav Maneriker, Kundan Krishna, and Natwar Modani.
\newblock 2018.
\newblock Corpus-based content construction.
\newblock In {\em COLING}.

\bibitem[\protect\citename{Strubell \bgroup et al.\egroup
  }2019]{strubell2019energy}
Emma Strubell, Ananya Ganesh, and Andrew McCallum.
\newblock 2019.
\newblock Energy and policy considerations for deep learning in nlp.
\newblock {\em arXiv preprint arXiv:1906.02243}.

\bibitem[\protect\citename{Vaswani \bgroup et al.\egroup
  }2017]{vaswani2017attention}
Ashish Vaswani, Noam Shazeer, Niki Parmar, Jakob Uszkoreit, Llion Jones,
  Aidan~N Gomez, {\L}ukasz Kaiser, and Illia Polosukhin.
\newblock 2017.
\newblock Attention is all you need.
\newblock In {\em NIPS}.

\bibitem[\protect\citename{Yang and Wang}2003]{yang2003fractal}
Christopher~C Yang and Fu~Lee Wang.
\newblock 2003.
\newblock Fractal summarization for mobile devices to access large documents on
  the web.
\newblock In {\em The Web Conference}.

\end{thebibliography}
\end{small}

\end{document}